\documentclass[sigconf]{acmart}

\usepackage{booktabs}
\usepackage{tabularx}
\usepackage{enumitem}
\usepackage{graphicx}
\usepackage{caption}
\usepackage{multicol} 
\usepackage[capitalize,noabbrev,nameinlink]{cleveref}
\usepackage{xspace}
\usepackage{placeins}
\usepackage{fancyhdr}

\usepackage{tikz}
\usepackage{eso-pic}
\AddToShipoutPictureBG{%
    \AtPageUpperLeft{%
        \raisebox{-1.5cm}{\hspace*{0.5\paperwidth}%
        \makebox[0pt][c]{\includegraphics[width=3cm]{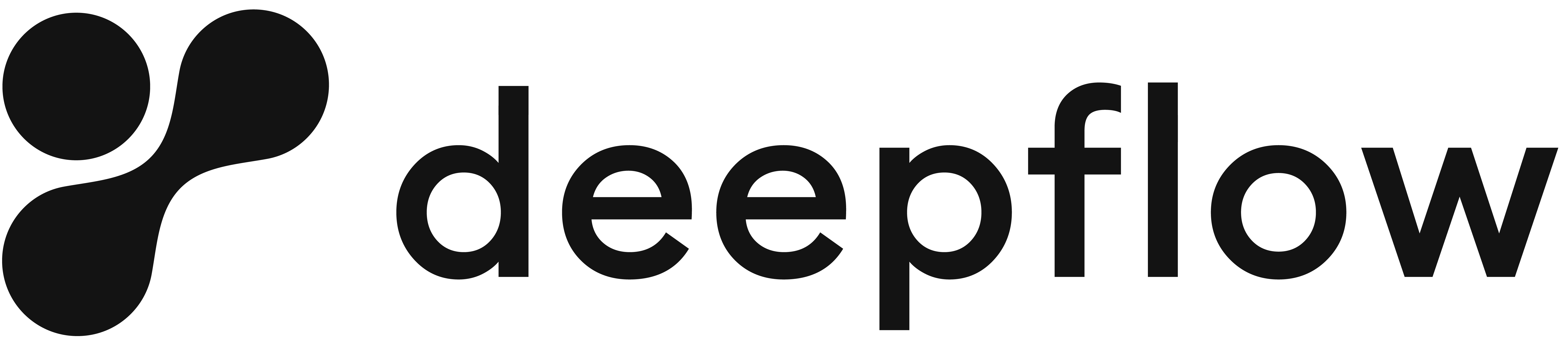}}}%
    }%
}

\usepackage{enumitem}
\setlist[itemize]{leftmargin=1.2em}
\setlist[enumerate]{leftmargin=1.5em}

\newcommand{\magym}{\textsc{\mbox{MA-Gym}}\xspace}

\AtBeginDocument{%
  }

\setcopyright{acmlicensed}
\copyrightyear{2025}
\acmYear{2025}
\acmDOI{XXXXXXX.XXXXXXX}

\acmConference[DAI '25]{Conference on Distributed Artificial Intelligence}{November 21--24, 2025}{London, UK}
\acmISBN{978-1-4503-XXXX-X/2025/11}

\begin{document}

\title{Orchestrating Human-AI Teams: The Manager Agent as a Unifying Research Challenge}

\author{Charlie Masters, Advaith Vellanki, Jiangbo Shangguan, Bart Kultys \\ Jonathan Gilmore, Alastair Moore, Stefano V. Albrecht}
\renewcommand{\shortauthors}{Masters et al.}
\affiliation{%
  \institution{DeepFlow}
  \city{London}
  \country{United Kingdom}
}
\email{{ charlie.masters, stefano.albrecht }@deepflow.com}

\begin{abstract}
While agentic AI has advanced in automating individual tasks, managing complex multi-agent workflows remains a challenging problem. This paper presents a research vision for autonomous agentic systems that orchestrate collaboration within dynamic human-AI teams. We propose the \textbf{Autonomous Manager Agent} as a core challenge: an agent that decomposes complex goals into task graphs, allocates tasks to human and AI workers, monitors progress, adapts to changing conditions, and maintains transparent stakeholder communication. We formalize workflow management as a Partially Observable Stochastic Game and identify four foundational challenges: (1) compositional reasoning for hierarchical decomposition, (2) multi-objective optimization under shifting preferences, (3) coordination and planning in ad hoc teams, and (4) governance and compliance by design. To advance this agenda, we release \magym, an open-source simulation and evaluation framework for multi-agent workflow orchestration. Evaluating GPT-5-based Manager Agents across 20 workflows, we find they struggle to jointly optimize for goal completion, constraint adherence, and workflow runtime--underscoring workflow management as a difficult open problem. We conclude with organizational and ethical implications of autonomous management systems.
\end{abstract}



\maketitle

\section{Introduction}

AI systems based on large language models (LLMs) have demonstrated remarkable proficiency at discrete, well-defined tasks across diverse domains, such as legal reasoning \cite{guha2023legalbench,hendrycks2021cuad}, software engineering \cite{jimenez2023swe}, drug discovery \cite{wu2018moleculenet,huang2022therapeutics}, and finance \cite{chen2021finqa,zhu2021tatqa}.

However, while current agentic systems are able to dynamically plan and execute well-scoped steps with high efficiency, the overarching strategic layer of workflow management consisting of task decomposition, dynamic resource allocation, progress monitoring, and adaptive re-planning remains fundamentally challenging \cite{xu2025theagentcompanybenchmarkingllmagents,liu2023agentbenchevaluatingllmsagents}.
This limitation becomes particularly pronounced in complex multi-agent environments where tasks are interdependent, objectives evolve dynamically, and coordination failures can cascade across entire workflows.

The next frontier in distributed AI lies in transcending task-level competence toward autonomous systems that can manage the entire life-cycle of complex, collaborative projects involving multiple stakeholders and evolving objectives. Our research vision is the creation of agentic ecosystems where autonomous AI agents and human workers collaborate as a team~\cite{vats2024survey}. To make progress toward this vision, we propose the {\bf Autonomous Manager Agent} as a specific research challenge.

\begin{figure*}[t]
    \includegraphics[width=\linewidth]{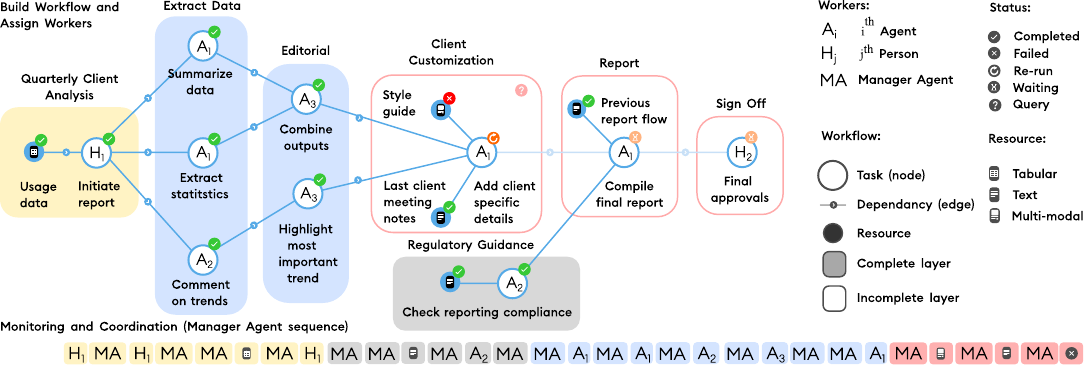}
    \caption{The Manager Agent (\texttt{MA}) as an orchestrator. Goal: ``Write an updated quarterly report for the client''. Based on this prompt, the \texttt{MA} is responsible for creating, modifying, and executing actions on a task graph, $G$, with a heterogeneous team of workers, $W$ (details in \cref{sec:formal}). The \texttt{MA} is responsible for coordinating the sequence of activity between workers, illustrated in the bottom row. Here, the \texttt{Client Customization} task does not complete as the resource fails verification, and the \texttt{MA} stops the execution, to then determine how to complete the flow successfully.}
    \label{fig:framework}
\end{figure*}

The Manager Agent is conceived as an autonomous entity responsible for the end-to-end management of complex workflows within a dynamic multi-agent environment. Its purpose is to orchestrate a team of human and AI ``workers'' to optimally achieve high-level goals specified by a human stakeholder. The responsibilities of the Manager Agent include decomposing complex goals into executable task graphs; allocating human and AI workers to tasks based on their skills, task requirements, and available resources; monitoring progress and proactively identifying potential impediments; adapting to changing environmental constraints and objectives; and maintaining transparent stakeholder communication. 
This vision aligns with the growing field of LLM-based Multi-Agent Systems (MAS), which aims to emulate the principles of human teamwork and specialization to solve problems collectively at scale~\cite{tran2025mas,yang2024llm}.

Realizing human-AI ecosystems demands coordinated research efforts across the Distributed AI community, spanning multiple traditionally separate sub-fields. Historically, scientific fields have often advanced by coalescing around ambitious challenge problems \cite{imagenet,bennett2007netflix,reddy1988foundations}.
The Manager Agent problem is designed to serve this role for distributed and agentic AI, requiring the integration of multiple sub-fields in multi-agent systems research, including multi-agent task decomposition and planning \cite{oliehoek2016concise}, multi-objective optimization and learning in dynamic teams with heterogeneous capabilities \cite{marl-book,mirsky2022survey}, and governance design \cite{yang2024llm}. The challenge is grounded in real-world workflow automation with scalable difficulty dimensions (e.g. number of tasks, team size, dependency complexity, constraints). By proposing this challenge, we aim to galvanize research efforts and provide a common ground for evaluating progress in creating collaborative agentic AI systems.

This paper makes the following contributions:
\begin{itemize}
    \item We propose the \textbf{Autonomous Manager Agent} as a unifying research challenge that bridges traditionally separate AI sub-fields including multi-agent coordination, compositional reasoning, and governance design (\cref{sec:framework}).

    \item We provide a \textbf{formal framework} modeling autonomous workflow management as a Partially Observable Stochastic Game (POSG), enabling principled algorithmic development (\cref{sec:formal}).

    \item We identify \textbf{four foundational research problems} for autonomous workflow management -- compositional reasoning for task decomposition, multi-objective optimization under non-stationary preferences, coordination in ad hoc teams, and governance by design (\cref{sec:challenges}).

    \item We release {\it Manager Agent Gym} (\magym): an \textbf{open-source library and simulator} for multi-agent workflow management (\cref{sec:ma-gym}).
    We evaluate Manager Agents based on GPT-5 across 20 diverse workflows and find that while they can achieve goal completion, constraint adherence, or workflow runtime, they fail to robustly solve for these qualities in tandem---highlighting workflow management as a difficult problem for agentic AI.

    \item We discuss the \textbf{ethical, regulatory, and privacy implications} of deploying autonomous management systems in real-world organizational contexts (\cref{sec:impacts}).

\end{itemize}

\section{Conceptual Framework: The Manager Agent}
\label{sec:framework}

The Manager Agent coordinates activity in a multi-agent system, analogous to a human project manager tasked with optimizing complex, large-scale workflows. 
Its environment is a dynamic task graph, illustrated in \cref{fig:framework}, where nodes represent tasks and directed edges represent dependencies. This graph is not static; it is created, modified, and executed by a heterogeneous team of ``workers,'' which can be specialized AI agents (e.g., a code-writing agent, a market-analysis agent) or human collaborators. This aligns with the MAS orchestrator-worker paradigm \cite{tran2025mas,krnjaic2024scalable}, where a coordinator decomposes problems and delegates sub-tasks to specialized agents. Thus, the Manager Agent focuses on high-level strategy and coordination while worker agents focus on task execution. Worker agents operate autonomously to execute their assigned tasks.

A key aspect of this framework is the shift from a ``human-in-the-loop'' model, where human intervention is required for most critical decisions, to a ``human-on-the-loop'' model: the human stakeholder retains control over high-level objectives and oversight, but the Manager Agent autonomously handles the intricate, moment-to-moment operational management. This paradigm shift promises to significantly amplify human productivity by offloading the cognitive burden of complex coordination, while keeping humans in a strategic, supervisory role.

\subsection{Core Capabilities}

To fulfil its role, the Manager Agent must possess a suite of core capabilities that mirror and extend those of a human manager:

\begin{enumerate}
\item \textbf{Structuring Workflows:} The Manager Agent must be able to take high-level, often ambiguous, goals from a human stakeholder and decompose them into a structured graph of feasible, concrete tasks and sub-tasks with clearly defined dependencies.

\item \textbf{Assigning Workers:} It must dynamically allocate tasks to the most suitable human or AI workers based on a deep understanding of task requirements, worker skills and availability, and resource constraints.

\item \textbf{Monitoring and Coordination:} The Manager Agent must continually track the progress of all workers on their assigned tasks, proactively identify and resolve bottlenecks, and ensure synchronized effort across the entire workflow.

\item \textbf{Adaptive Planning and Execution:} The environment is dynamic. The Manager Agent must be able to generate and execute modifications to the workflow in real-time---revising the task graph, adjusting worker roles, or reassigning tasks in response to unexpected events, information, or changing priorities.

\item \textbf{Stakeholder Communication:} It must maintain transparent communication with the stakeholder, providing regular updates on plans, progress, and potential issues to ensure robust oversight and enable informed intervention when necessary.
\end{enumerate}

\subsection{A Unifying Challenge}

The Manager Agent problem is not merely a practical application, but a unifying research challenge that necessitates a synthesis of capabilities from traditionally separate sub-communities within multi-agent systems (MAS) research. Its core capabilities are directly related to long-standing research questions in task decomposition and allocation~\cite{khamis2015multi}, team formation~\cite{liemhetcharat2012modeling}, multi-agent coordination and learning~\cite{marl-book}, communication~\cite{zhu2024survey}, agent modeling~\cite{albrecht2018modelling}, and ad hoc teamwork~\cite{mirsky2022survey}.

Specifically, the capabilities of \textbf{Structuring Workflows} and \textbf{Assigning Workers} are a direct instantiation of the central MAS problem of task decomposition and allocation, which seeks optimal methods for breaking down complex goals and mapping sub-tasks to the most appropriate agents~\cite{khamis2015multi}. This process is also intrinsically linked to team formation \cite{liemhetcharat2012modeling}, as the Manager Agent must dynamically assemble a group of workers to execute the workflow. The capabilities of \textbf{Monitoring and Coordination} and \textbf{Adaptive Planning and Execution} are the essence of multi-agent coordination and collaboration, a field focused on managing dependencies and ensuring coherent collective action~\cite{oliehoek2016concise}. To improve over time, the Manager Agent must engage in multi-agent learning to refine its strategies for decomposition, allocation, and coordination~\cite{marl-book}, as well as agent modeling \cite{albrecht2018modelling,nashed2022survey} to infer the capabilities and state of its workers. Effective communication is vital, not only for \textbf{Stakeholder Communication} but also for the coordination protocols between the manager and workers, a topic that includes the potential for learning emergent languages~\cite{zhu2024survey}. Because the team composition is not fixed, this entire process must occur under the challenging conditions of ad hoc teamwork \cite{mirsky2022survey,JRahman2022POGPL}, where the Manager Agent must be able to collaborate with new teammates without pre-coordination between agents (such as prior joint training).

The ambition to build such a unifying agent is not new, but its feasibility has been unlocked by recent breakthroughs in high-capacity foundation models~\cite{bommasani2021opportunities, comanici2025gemini25pushingfrontier}. These models, particularly Large Language Models (LLMs), provide the ``cognitive engine'' for the Manager Agent, capable of high-level reasoning and planning across a range of complex, real-world tasks that was previously intractable for automated systems~\cite{aghzal2025survey}. The emergence of Large Reasoning Models (LRMs) in 2024-2025 marks a significant milestone~\cite{openai2024o1,deepseekai2025deepseekr1incentivizingreasoningcapability}. These models leverage large-scale reinforcement learning to achieve stepwise reasoning required for dynamic planning and adaptation in complex workflows~\cite{deepseekai2025deepseekr1incentivizingreasoningcapability}. This creates a unique opportunity to synthesize the reasoning power of foundation models with established MAS frameworks, positioning the Manager Agent as a timely and achievable research goal.

\section{A Formal Model of Workflow Management}
\label{sec:formal}

To ground this challenge in a mathematical framework, we model the problem of autonomous workflow management as a \textbf{Partially Observable Stochastic Game (POSG)} \cite{hansen2004dynamic}. A POSG models scenarios where multiple agents interact in a shared environment with incomplete information and different objectives. This is an appropriate model for our domain, as it explicitly accounts for the Manager Agent and the team of worker agents as distinct decision makers with their own action sets, observations, and preferences.
A POSG is defined by the tuple $\langle I,S,b^0,\{A_i\},\{O_i\},P,\{R_i\} \rangle$, where each component is specified for our domain as follows.

\subsection{Set of Agents ($I$)}

The set of agents $I$ consists of the Manager Agent ($M$) and the set of all worker agents ($W$), which can include both human and AI workers. Thus, $I=\{M\}\cup W$. (We will remark on modeling the stakeholder in \cref{sec:stakeholder}.)

\subsection{State Space ($S$)}
\label{sec:posg-state}

The underlying state of the environment at any time $t$, denoted $s^t\in S$, is a comprehensive snapshot of the entire workflow. It is defined as a tuple $s\equiv\langle G,W,C,X,U\rangle$:

\begin{itemize}
    \item $G$: The complete task-dependency graph, including all task nodes $\{T\}$, their metadata $\mu_{T}$ (e.g. status, owner, progress to date) and directed edges $E_{T_i, T_j}$ representing dependencies between tasks.%
    \item $W$: The set of all available human and AI workers, including their capabilities, current assignments, availability, and cost rates.
    \item $C$: A persistent set of communications $\{C_{i,j}\}$ between all agents $i,j \in I$.
    \item $X$: A registry of all artifacts produced by tasks so far, including documents, code, and other digital assets.
    \item $U$: A set of preference weights of the stakeholder for how tasks are completed (e.g. cost, speed, quality). These may be fully or partially observable to agents $I$, and can evolve over time.
\end{itemize}
The initial state $s^0$ is sampled from the initial state distribution $b^0$.

\subsection{Action Spaces ($A_i$)}

Each agent $i\in I$ has its own set of available actions it can perform.

\textbf{Manager Agent's Action Space ($A_m$):} The Manager Agent possesses a rich action space designed to orchestrate complex workflows through three primary categories of capabilities:

\begin{itemize}
\item \emph{Observability-increasing actions:} These actions allow the agent to reduce its uncertainty about the state by gathering information about the workflow's progress. Representative examples include \texttt{Inspect($T_i$)} to view execution logs and outputs $\mu_{T_i}, X_{T_i}$ for specific tasks, and \texttt{GetChatHistory($T_i$)} to retrieve the recent communication history related to a given task.%

\item \emph{Graph-modifying actions:} These actions dynamically alter the structure of the workflow itself. Representative actions include \texttt{AddTask}($T_i$) and \texttt{RemoveTask}($T_i$) to add or remove tasks $T_i$ from $G$, \texttt{AddEdge($T_i, T_j$)} to establish or modify dependencies $E_{T_i,T_j}$ between tasks, and \texttt{DecomposeTask($T_i$)} to decompose a complex task into sub-tasks. %

\item \emph{Delegation and communication actions:} These actions manage the team of workers. Examples include \texttt{AssignTask($T_i, W_j$)} to assign a task to a human or AI worker, and \texttt{SendMessage($W_j$,message)} to communicate directly with a worker to provide guidance, request updates, or alert them to changing priorities or constraints.
\end{itemize}

\textbf{Worker Agent's Action Space ($A_i$ for $i\in W$):} Worker agents have an action space focused on task completion, including a set of tools which may be fixed or dynamic, or the ability to seek additional information when task requirements are ambiguous.

The evolution of the system depends on the \textbf{joint action} $a^t=\langle a^t_1,\ldots,a^t_n\rangle\in\times_{i\in I}A_i$ taken by all agents at time $t$.

\subsection{Observation Spaces ($O_i$)}

Each agent has a private, partial view of the state, as well as (potentially) the past actions of other agents.

\textbf{Manager Agent's Observation Space ($O_M$):} The Manager Agent has full access to some parts of the state, such as the high-level task graph, resource metadata, and chat history, but may not directly observe other parts such as stakeholder preferences, or the contents of artifacts or detailed worker logs.

\textbf{Worker Agent's Observation Space ($O_i$ for $i\in W$):} A worker's observation may be limited to the details of its assigned tasks, communications sent to or by the worker, and artifacts generated by the worker. Additional information about the workflow state may be observable to workers depending on their roles within the team.

\subsection{Transition and Observation Dynamics ($P$)}
\label{sec:posg-trans}

The function $P(s',o|s,a)$ defines the probability of transitioning to state $s'$ and all agents receiving a joint observation $o=\langle o_1,\ldots,o_n\rangle$, given the current state $s$ and the joint action $a$ taken by all agents.
The state dynamics can be deterministic, such as actions that modify the task graph (e.g. removing a task). State dynamics can also be stochastic, particularly when actions involve workers whose performance is non-deterministic (e.g. time to task completion, quality of task output, etc.).

\subsection{Reward Functions ($R_i$)}

Each agent $i\in I$ has its own reward function $r_i = R_i(s,a)$, reflecting its preferences and objectives. We also include workflow-level constraints: hard constraints $\mathcal{H}$ that must always hold, and soft constraints $\mathcal{S}$ that can be violated with penalties.

\textbf{Manager Agent's Reward ($R_M$):} The Manager Agent's reward is aligned with the high-level goals of the stakeholder, represented by its preferences $U$. Conceptually, this is a multi-level function that includes a sparse terminal reward for successful workflow completion, and can include additional performance metrics such as time and cost for workflow execution and overall quality of work outputs, weighted according to stakeholder preferences $U$. Additionally, violations of hard and soft constraints ($\mathcal{H}/\mathcal{S}$) result in penalties of varying magnitudes (e.g. large penalty for violating hard constraints).

\textbf{Worker Agent's Reward ($R_i$ for $i \in W$):} A worker agent's reward may be simpler, based on metrics such as the timely and successful completion of its assigned tasks. This allows for modeling self-interested behavior, where a worker might prioritize its individual task goals over the global workflow objective.

If all agents share the same reward function, i.e. $R_i = R, \forall i \in I$, then the model simplifies to a Decentralized POMDP (Dec-POMDP), which is a purely cooperative setting \cite{oliehoek2016concise}. The more general POSG formulation allows for a richer spectrum of mixed cooperative and self-interested behaviors.

\subsection{Solution Concept}

The actions of each agent $i \in I$ in the system are governed by a policy $\pi_i$, which assigns probabilities to the agent's available actions $A_i$ based on the history of the agent's observations $\{o_i^t\}_{t = 0,1,2,...}$.

The {\bf optimal policy} $\pi_M^*$ of the Manager Agent will depend on the assumptions we make about the behaviors of other agents $W$ in the system. Borrowing concepts from Stochastic Bayesian Games \cite{albrecht2016belief} and Interactive POMDPs \cite{doshi2006difficulty}, we may assume that the worker agents $j \in W$ draw their policies from a set of possible policies $\pi_j \in \Pi_W$. In this case, we seek a policy $\pi_M^*$ for the Manager Agent that maximizes the expected return, given by the expected sum of rewards $r_M^t$ it receives over time $t=0,1,2,...$, assuming the worker agents can use any of the policies in $\Pi_W$. Such an assumption may be suitable if the possible behaviors of workers are well understood, which is often the case for defined AI workers.

In general, we may also assume that some worker agents may learn and adapt their behaviors over time based on past observations, as is the case with human workers and more advanced AI workers. This brings us into the realm of game theory and multi-agent reinforcement learning (MARL) \cite{marl-book}. In a POSG, where agents may have different preferences, the optimal solution becomes an \textit{equilibrium}---a joint policy from which no agent has a unilateral incentive to deviate. This idea is embodied by the \textbf{Nash Equilibrium (NE)} \cite{nash1950equilibrium}, which is a joint policy $\pi^*=(\pi_1^*,\ldots,\pi_n^*)$ in which for every agent $i$, its policy $\pi_i^*$ is a best response to the set of other agents' policies $\pi^*_{-i} = \pi^* \setminus \{\pi^*_i\}$. This means that no agent $i$ can improve its expected return by changing its policy alone. Furthermore, a joint policy $\pi^*$ is \textbf{Pareto-optimal} if there is no other joint policy $\pi'$ that can increase the expected return for at least one agent without decreasing the expected return for any other agent.

Combining these concepts, one solution concept for our POSG is a \textbf{Pareto-optimal Nash Equilibrium (PONE)}~\cite{munoz2006learning}. This is a joint policy that is both stable (a Nash Equilibrium) and efficient (Pareto-optimal), representing a desirable outcome for the collaborative human-AI team. In the Dec-POMDP case, in which agents share the same rewards, this corresponds to a joint policy that maximizes the expected return of all agents.

\subsection{Modeling the Stakeholder}
\label{sec:stakeholder}

There are different ways to model the stakeholder in our POSG formalism. In the most general case, a stakeholder is itself an autonomous agent that can observe partial information about the workflow state, and take actions to interact with the workflow and other agents (manager, workers). Thus, a stakeholder agent $\alpha \in I$ can be represented via its own action set $A_\alpha$, observation set $O_\alpha$, and reward function $R_\alpha$. The stakeholder's action set may include those of the Manager Agent and additional actions to enable communication with the Manager Agent and updating stakeholder preferences $U$ (\cref{sec:posg-state}).

If the stakeholder is passive and does not directly choose actions in the POSG, thus giving full control to the Manager Agent, we may instead model the stakeholder as part of the transition dynamics $P$ (\cref{sec:posg-trans}). By including the time index $t$ inside state $s^t$, the stakeholder can change its preferences $U \in s^t$ over time $t$ through state transitions $P(s^{t+1},o^{t+1} | s^t,a^t)$.

\section{Foundational Research Challenges}
\label{sec:challenges}

The realization of an autonomous Manager Agent requires significant advances across several core areas of AI. We identify four foundational research challenges that are critical to our vision.%

\subsection{Hierarchical Task Decomposition}\label{sec:decomposition}

For a Manager Agent, mapping a high-level workflow description into a task graph $G$ with governance constraints $\{\mathcal{H},\mathcal{S}\}$ is the bottleneck that unlocks all downstream capabilities (e.g. allocation, monitoring, adaptation).
Recent LLM-based multi-agent frameworks \cite{bai2024twostep,yu2025dyntaskmas} show that performance gains correlate almost linearly with the quality of the induced task graph—underlining that structure learning, not raw language generation, is the critical path.
Empirical studies reveal that both single-agent LRMs and multi-agent orchestration systems fail once graph depth, branching factor or novelty exceed modest thresholds. For example, large-scale audits of reasoning tasks~\cite{kwa2025measuringaiabilitycomplete,lin2025zebralogicscalinglimitsllms} find sharp phase transitions beyond which success probability collapses. Multi-agent variants inherit these limits, with recent research \cite{wang2025tdag,wang2025megaagent} reporting substantial error cascades when sub-agents produce incompatible sub-plans, despite significant engineering for coordination.
Current agents can discover local patterns but lack a hierarchical notion of causality that scales with problem complexity, prompting the question:

\textit{How can a Manager Agent scale to robustly solving large complex planning problems in dynamic multi-agent systems?}

Why might models struggle? Recent work shows that transformer-based agents appear to rely on shallow subgraph matching over memorized patterns rather than genuine composition~\cite{dziri2023faithfatelimitstransformers}.  This shortcut is effective for in-distribution tasks but can fail when unseen combinations or long-range dependencies dominate in dynamic, heterogeneous teams. Moreover, partial observability and non-stationary interactions make these problems worse in multi-agent settings. They amplify exposure bias: when one worker deviates locally, errors propagate through the task graph. This invalidates the Manager Agent's original planning assumptions made to satisfy environment constraints $\mathcal{H}$, $\mathcal{S}$ and preferences $ U$.

Despite this, we can find some promising directions to tackle these problems in the following areas:

\begin{enumerate}[nosep,leftmargin=*]
  \item \textbf{Structured latent planning.}  Augment LRMs with explicit symbolic planners or graph-structured controllers that operate over learned abstractions, enabling verifiable sub-plan composition in the style of neuro-symbolic planning~\cite{capitanelli_2024,liu2023llmp,besta2024graph}.
  \item \textbf{Meta-adaptive decomposition.}  Treat task-graph induction itself as a meta-RL problem: train the manager to iteratively propose, simulate and revise decompositions with self-consistency and outcome feedback~\cite{lambert2025tulu3pushingfrontiers}.
\end{enumerate}

\subsection{Multi-Objective Optimization with Non-Stationary Preferences}

An optimal policy $\pi_M^*$ for the Manager Agent must juggle multiple, often competing objectives such as \emph{cost}, \emph{latency} and \emph{quality}, yet the workflow stakeholder can re-rank these objectives mid-execution.  This creates a \textbf{dynamic multi-objective} problem layered on top of the usual multi-agent coordination issues.

Two separate sub-fields attack aspects of this problem, but neither address the entire setting. Multi-Objective Reinforcement Learning (MORL) provides a framework for optimizing multiple, potentially conflicting objectives via \textbf{scalarization methods} that combine objectives via weighted sums~\cite{morl_scalar}, and \textbf{Pareto-based methods} that learn policy sets representing different trade-offs~\cite{vanmoffaert2014multi}; both of which assume the objectives are fixed a priori, and break when preference weights shift online. 

Conversely, modern preference-learning pipelines model \emph{one} scalar reward and thus ignore multi-objective trade-offs. In reinforcement learning from human feedback (RLHF), a reward model is trained to score an output higher than a ranked alternative, and the policy is fine-tuned to maximize that score \cite{rafailov2024directpreferenceoptimizationlanguage}. In contrast, reinforcement learning from verifiable rewards (RLVR) replaces this learned model with a verifiable reward, modeling preference in task execution via reward shaping around easily measured attributes such as length or clarity.
\cite{lambert2025tulu3pushingfrontiers,deepseekai2025deepseekr1incentivizingreasoningcapability}.  Both lines of work assume the reward remains fixed; recent extensions handle \emph{non-stationary} scalar preferences \cite{son2025right} but still optimize a single objective—an expressivity gap formalized in \cite{abel2022expressivitymarkovreward}.

Since an effective manager must pragmatically handle evolving, conflicting objectives, the open research question remains:

\emph{How can a Manager Agent learn a robust policy that can efficiently adapt to non-stationary stakeholder preferences over multiple objectives without requiring costly retraining?}

Early answers may lie in test-time alignment and meta-learning that infer fresh weight vectors from a few stakeholder corrections \cite{xu2025genarm,nguyen2025multiattributesteeringlanguagemodels}, or explicit hierarchical control where the Manager Agent re-weights objectives based on stakeholder interactions while task agents solve the resulting single-objective sub-tasks, both of which remain unexplored in verifiable-reward, multi-agent settings.

\subsection{Coordination in Ad Hoc Teams}

The Manager Agent must orchestrate collaboration in dynamic, heterogeneous teams where agents may join or leave without prior coordination. This means the Manager Agent cannot rely on pre-coordinated strategies or prior joint training with a fixed set of teammates. This defines the classic \textit{ad hoc teamwork} (AHT) problem---a long-standing challenge in multi-agent systems \cite{stone2010ad,mirsky2022survey}.

Key open problems in AHT that are directly relevant to our setting include generalizing to new types of teammates with their individual capabilities, expectations, and working preferences, as well as effectively collaborating with teammates who are themselves learning and adapting their behavior~\cite{mirsky2022survey}. The Manager Agent must be able to quickly infer the skills, knowledge, and preferences of workers based on limited interactions, and flexibly adapt how it communicates and coordinates with workers \cite{albrecht2016belief,barrett2015making}.
The open research question is therefore:

\emph{How can the Manager Agent rapidly infer the capabilities, reliability, and intent of new teammates from limited interaction and leverage this understanding for effective, on-the-fly task delegation and coordination?}

Several approaches point toward partial solutions. Ribeiro et al.~\cite{ribeiro2024teamster} use model-based reinforcement learning to learn teammate behavior and adapt policies on the fly, but their method assumes sufficient prior interaction and struggles under extreme heterogeneity. Zhang et al.~\cite{zhang2025taget} propose training on offline trajectories to predict teammate-aware goals, though their approach lacks the ability to reason about unobserved agent types. Wang et al.~\cite{wang2024nagent} embed teammate behaviors to support policy generalization in novel teams, but still assume consistent observation structures and offer limited mechanisms for dynamic role negotiation. Liu et al.~\cite{liu2024heterogeneous} leverage large language models for hierarchical plan generation using interactive reasoning, though their reliance on language abstractions may not scale to low-level execution. Jin et al.~\cite{jin2023capability} introduce a capability-aware ad hoc teamwork model using agent hierarchies, but this approach presumes known capability classes.

Each method tackles a facet of the problem: policy adaptation, team modeling, or high-level coordination, but none yet provide the full-stack reasoning, real-time inference, and dynamic task restructuring required for Manager Agents to operate robustly in open, evolving teams. Ad hoc coordination remains a critical and unsolved challenge for collaborative AI.

\subsection{Governance and Compliance by Design}

Manager Agent autonomy in complex organizational workflows creates a critical challenge: maintaining governance and compliance across dynamic multi-agent systems. These agents must interpret natural language constraints, adapt to evolving regulations, and demonstrate compliance across heterogeneous teams. The rapid advancement and deployment of AI to safety critical environments poses significant regulatory challenges with highly unpredictable and rapidly changing requirements~\cite{dimitriou2024ai,kuznietsov2024avreview}.
Solving these challenges involves tackling a series of problems, specifically:

\emph{How can Manager Agents maintain governance and compliance in dynamic multi-agent workflows while adapting to evolving regulatory constraints without compromising operational effectiveness?}

Multi-agent constraint satisfaction requires ensuring workflow-level compliance across dynamically changing teams where agents have heterogeneous capabilities and roles. Recent works attempt to address distributed safety coordination. Gu et al.~\cite{gu2024scalable} make progress on safe decentralized MARL via scalable constrained policy optimization, but only in a static team composition setting. Aydeniz et al.~\cite{aydeniz2025safe} achieve team-level constraint satisfaction through joint entropy maximization but only for binary collision avoidance scenarios, not tackling the complex and ambiguous requirements found in regulatory compliance. 

The Manager Agent's need to interpret natural language constraints demands translating ambiguous regulatory text into executable policies that can guide agent behavior. Existing approaches address binary safety constraints versus nuanced regulatory requirements that are inherently ambiguous, where optimal policies depend on how organizations choose to handle uncertainty and interpretation. Yao et al.~\cite{yao2023constraint} achieve zero-shot adaptation to varying constraint thresholds using conditioned policy optimization, but do not address the problem of mapping from complex regulatory reasoning, focusing purely on numerical parameter adjustment.

Real-time governance adaptation requires adapting to varying safety constraints during deployment without retraining, as regulatory landscapes change post-deployment while agents must continue operating. Recent interpretability advances attempt to enable runtime safety analysis. Anthropic researchers~\cite{anthropic2025attribution, anthropic2025tracing} achieve training-time safety analysis through mechanistic interpretability and circuit tracing, but do not address automated post-deployment adaptation to regulatory changes. This connects to the broader challenge that test-time alignment remains an open problem, and optimal helpfulness-harmlessness tradeoffs are domain-specific, often creating conflicts between compliance objectives \cite{bai2022constitutionalaiharmlessnessai,ganguli2022redteaminglanguagemodels}.

Some promising solution areas to these problems could be found in \textit{ad-hoc constraint-aware teaming} which extends existing team constraint approaches~\cite{gu2024scalable, aydeniz2025safe} to dynamic compositions, \textit{natural language constraint grounding} that combines existing work into using control barrier function learning with LLM-based regulatory interpretation~\cite{yao2023constraint}, and further mechanistic research to better understand the underlying traces of models inner workings \cite{gyevnar2024cema}.

\section{Manager Agent Gym: A Simulator for Human--AI Workflow Orchestration}
\label{sec:ma-gym}

Progress on the four challenges discussed in \cref{sec:challenges} demands a single testbed that exercises hierarchical control, dynamic multi-objective preferences, ad-hoc teaming, and governance \emph{together}. As summarized by the comparison in \cref{tab:benchmarks} (\cref{app:benchmarks}), existing benchmarks each cover some of these aspects but none evaluates the full spectrum of Manager Agent capabilities.

To fill this gap, we release \textbf{Manager Agent Gym} (\magym):\footnote{\magym codebase: \url{https://github.com/DeepFlow-research/manager_agent_gym}} a discrete-timestep environment where a manager operates over a graph-based multi-agent workflow and must address any or all of the four challenges per episode. We provide a high-level description of \magym in \cref{sec:ma-gym-spec} along with initial benchmark results in \cref{sec:results}. Detailed specifications can be found in \cref{app:magym} and the \magym code repository.

\subsection{\magym Overview and Workflows}
\label{sec:ma-gym-spec}

\magym instantiates the POSG formalism in \cref{sec:formal} with an initial task dependency graph $G$ and agent team $I$ consisting of AI workers and simulated human workers, which are able to communicate via actions that invoke a communication store $C$. Additionally, \magym includes a stakeholder agent $\alpha$ (\cref{sec:stakeholder}) with preference weights $U$, which can choose to take actions based on its policy $\pi_\alpha$ configured in the simulator, including communicating with other agents, updating stakeholder preferences, and answering questions from the Manager Agent.

\begin{figure*}[!htbp]
    \centering
    \includegraphics[width=1.0\linewidth]{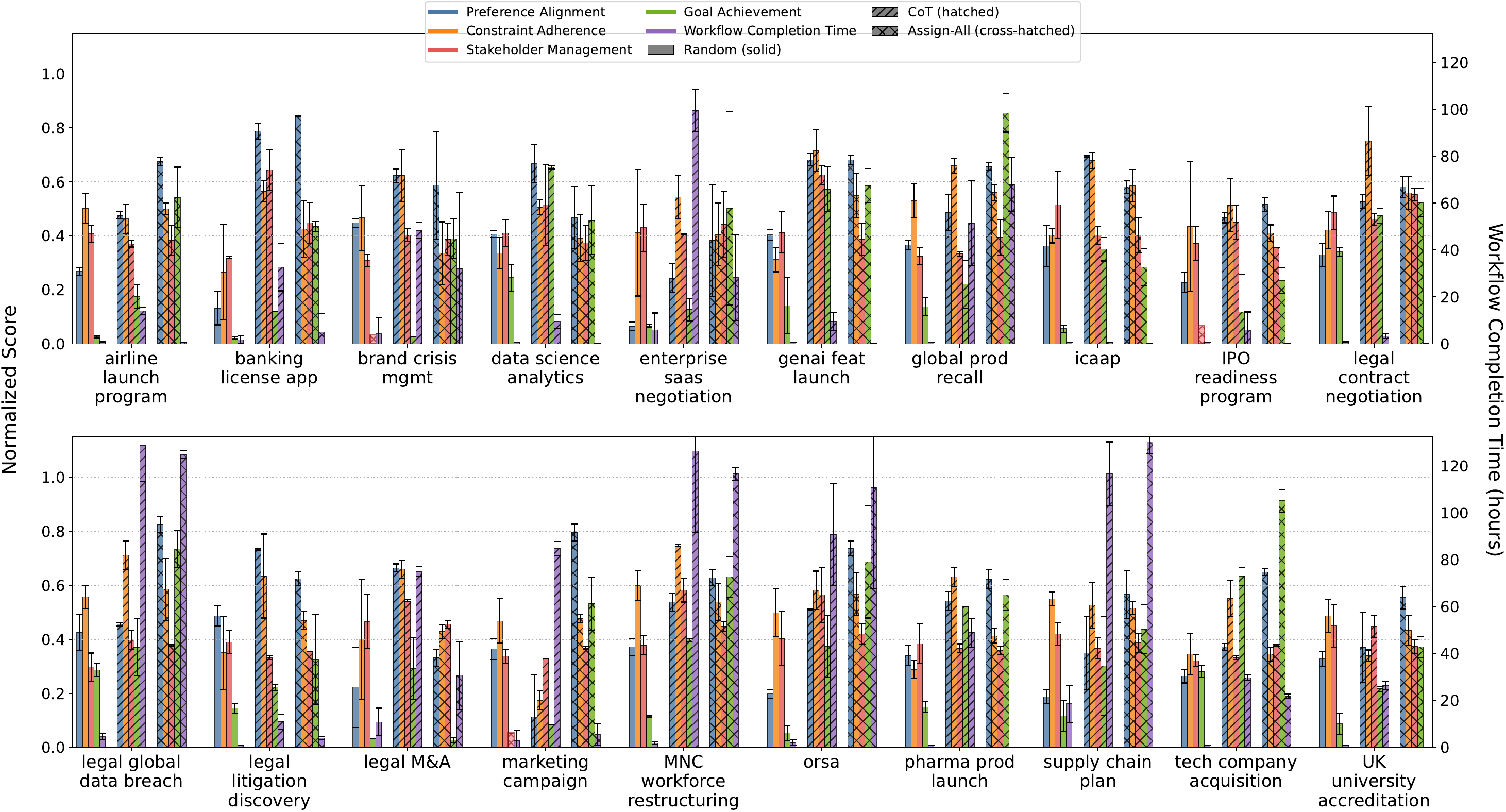}
    \caption{Random, Chain-of-Thought (CoT) and Assign-All policy performances plotted across 20 workflows (bars show average and standard deviation across 5 random seeds per workflow). Details of workflows can be found in \cref{app:taxonomy}, \cref{tab:workflows}.}
    \label{fig:results}
\end{figure*}

\magym runs each workflow episode as defined in our POSG model:
at each timestep, each agent $i \in I$ (including Manager and stakeholder agents) gets an observation and takes an action from its set of available actions $A_i$ (full action sets in \cref{app:manager-actions}, \cref{tab:manager-actions-rationale}) based on its policy $\pi_i$. The actions are then executed in \magym, and all tasks that are ready and have been assigned to a worker are executed. Worker agents can join and leave the team at any timestep during an episode, as configured in \magym.

As part of the initial release of \magym, we define 20 challenging workflow scenarios representing a diverse range of real-world domains, with different stakeholder preferences, graph complexity (number of nodes and dependencies), team composition (number and types of agents), and hard/soft constraints. A full listing of these workflows can be found in \cref{app:taxonomy}, \cref{tab:workflows}.

\subsection{Baselines and Results}
\label{sec:results}

We evaluate three baselines of LLM-based Manager Agents with OpenAI GPT-5 as a base model. \textbf{Random:} observes the dependency graph $G$ and at each timestep is restricted to taking an action chosen uniform-randomly from $A_M$ (the agent still has to specify input parameters for the sampled action); \textbf{CoT:} observes the dependency graph and uses Chain-of-Thought reasoning to choose its next action from the full action set $A_M$; \textbf{Assign-All:} reads the starting state of the workflow $G$ and assigns each task in $T$ to the most suitable human or AI worker based on their skill descriptions and task requirements, thus performing all workflow planning upfront. A list of allowed Manager Agent actions can be found in \cref{app:manager-actions}. We repeat all 20 workflows across 5 random seeds, capping the maximum number of Manager Agent actions at 100 before terminating the episode. We report on Preference alignment, Constraint adherence, Goal achievement, Stakeholder management, and Workflow completion time in \cref{fig:results}; see \cref{app:metric-defs} for definitions of these metrics.

\textbf{How challenging is \magym as an environment?}
Across the 20 workflows, all baselines struggle to balance goal completion, completion time, and constraint adherence. On average, CoT achieves only modest goal completion ($0.313 \pm 0.187$), a limited uplift over Random ($0.135 \pm 0.098$). The Assign-All baseline, despite lacking adaptive planning or oversight, achieves higher goal completion ($0.502 \pm 0.209$), suggesting that managerial interventions can sometimes be actively detrimental. Yet this advantage is fragile: Assign-All shows lower constraint adherence ($0.475 \pm 0.080$) compared to CoT ($0.589 \pm 0.140$), only marginally higher than Random ($0.432 \pm 0.095$). Grouping workflows by capability demands reveals systematic variance in our objectives across baselines: Action-heavy processes (e.g., airline launches, marketing campaigns, AI product rollouts) strongly favor Assign-All, which achieves on average 0.373 higher goal completion compared to CoT. In documentation and audit-heavy workflows (e.g., contract negotiation, university accreditation), and CoT retains a clear lead in constraint adherence (0.579 vs.\ 0.419).
These mixed outcomes emphasize that no single baseline performs consistently well across domains.

\textbf{What trade-offs do we observe in different policies?}
The differences above reflect a series of trade-offs in the \magym setting. CoT reliably completes most generated task nodes (80\% vs.\ 0\% for Random), but at great cost: average runtime rises to 46.9 hours compared to 2.7 for Random, with 25.8\% delegation overhead and end-to-end execution 17$\times$ slower. These slowdowns stem from dependency bottlenecks, where human actions block downstream tasks. Assign-All reduces such stalls by dispatching tasks in bulk, cutting runtime dramatically in 16 of 20 workflows (e.g., $-82.9$ hours in Marketing; $-68.3$ hours in SaaS) and achieving higher average goal completion than CoT. Yet these gains come only by bypassing reasoning and sequencing, which weakens constraint adherence and stakeholder engagement. Taken together, this reveals a multidimensional trade-off space: goal achievement, completion time, constraint adherence, and stakeholder engagement cannot all be maximized simultaneously.

We also study the efficacy of ``reasoning models'' trained with verifiable rewards in place of ``traditional'' LLMs, comparing GPT-5 and GPT-4.1 under identical conditions. We find that GPT-5 achieved consistently higher goal completion (0.6--0.7 on analytics and product-launch workflows; see \cref{fig:gpt41-tradeoff}) and deployed richer planning operators, executing 14$\times$ more decompositions and 26$\times$ more dependency links than GPT-4.1 (see \cref{tab:action-freq}). GPT-4.1 instead relied heavily on messaging and status queries, resembling a reactive communicator (see \cref{app:gpt41} for further discussion). This highlights that stronger reasoning supports more proactive orchestration but does not eliminate the bottlenecks, stakeholder neglect, or constraint violations.

\section{Organizational and Ethical Considerations}
\label{sec:impacts}

The deployment of an autonomous Manager Agent, a system with the authority to delegate tasks and manage resources for a team of humans and AIs, raises significant organizational and ethical considerations. A responsible research program must proactively address these ethical, regulatory, and privacy challenges, which are central to the development of trustworthy AI \cite{kaur2022trustworthy,kuznietsov2024avreview}. 

Considerations arise at the level of an individual task assigned to a worker, at the level of the workflow, and as a result of the changes to workflows over time. For example, a failure in an agent-managed workflow may result in the unjust attribution of blame to a human worker assigned a specific task. This scenario exemplifies the phenomenon known as the 'moral crumple zone', where responsibility for systemic failures is placed disproportionately on human operators, while the underlying complexity and opacity of the technological system obscure the accountability of other actors, such as designers, developers, and institutional stakeholders \cite{elish2016moral}. 

A well-designed Manager Agent, however, has the potential to mitigate these risks. The agent's architecture can be designed for transparency. For example, the communication store ($C$) and the detailed action logs can serve as an immutable audit trail. Actions like \texttt{SendMessage} and \texttt{Inspect} are not just for coordination, but are crucial mechanisms for maintaining human oversight and creating shared context. By making its reasoning process explicit and traceable, the agent can help construct clear accountability frameworks where responsibility is distributed fairly among all participants, human and AI, based on their actual control and contribution~\cite{casalicchio2022accountability}.

The Manager Agent's core function of assigning tasks to workers is a form of automated resource allocation. Without careful design, this process is susceptible to bias, which could perpetuate or amplify existing inequities~\cite{mehrabi2021survey}. For example, the agent might learn spurious correlations from historical data and consistently assign more desirable, high-growth tasks to one group of workers (e.g., AI agents or a specific human demographic) while relegating another group to tedious, low-impact work~\cite{eve2025common}. There will be scenarios where there is value in humans performing tasks to facilitate learning, reduce skills atrophy or promote social cohesion. 

This challenge connects to the extensive body of research on fairness in AI and resource allocation~\cite{chouldechova2017fair}. To address this, fairness cannot be an afterthought. We propose that formal fairness criteria---such as those based on principles of envy-freeness or maximin share~\cite{budish2011combinatorial}---must be integrated directly into the agent's objective function. By making fairness an explicit optimization goal alongside efficiency and quality, we can guide the agent's learning process toward equitable and just allocation policies.

Effective management requires monitoring, but monitoring creates significant privacy risks, particularly for human workers whose activities, communications, and performance are being tracked~\cite{eve2025common}. The agent's access to the communication history ($C$) and task artifacts ($X$) could expose sensitive personal or proprietary information.
This necessitates research into privacy-preserving architectures \cite{shi2025privacy}.
For instance, the agent might use federated learning to learn from worker data without centralizing it, or employ differential privacy to gather aggregate statistics on team performance without revealing individual contributions~\cite{mcmahan2017communication}. A key challenge is that different agents (e.g., internal employees vs. external contractors) may have heterogeneous privacy requirements, demanding flexible and context-aware privacy protocols~\cite{shi2025privacy}. The goal is to design a system that achieves the necessary level of observability for effective management while protecting the privacy of all participants.

Ultimately, the safe and responsible deployment of a Manager Agent requires a comprehensive governance framework. This framework must synthesize the elements discussed above, establishing clear ethical principles, mandating transparency and explainability (XAI) mechanisms~\cite{adadi2018peeking}, and implementing robust protocols for continuous monitoring and auditing.

\section{Conclusion and Future Work}
\label{sec:conclusion}

This paper has outlined an ambitious vision: an autonomous Manager Agent that orchestrates dynamic teams of human and AI agents to solve complex problems. This timely research goal, enabled by advances in large language and reasoning models, integrates multiple areas of multi-agent systems research. We formalized autonomous workflow management as a structured POSG and identified four core technical challenges: compositional reasoning, multi-objective optimization with non-stationary preferences, ad hoc team coordination, and governance by design. The Manager Agent problem unifies these disparate research threads into a holistic challenge.

To support research on the Manager Agent challenge, we have released the Manager Agent Gym (\magym) which implements the POSG formalism and supports algorithm design and evaluation for Manager Agents. We benchmarked LLM-based Manager Agents across a diverse set of workflows inspired by real-world tasks, showing that jointly optimizing for goal achievement, constraint adherence, and resource usage (e.g. workflow runtime) is a difficult problem for agentic AI. Our future work includes building on this initial release of \magym by defining additional challenging workflow scenarios, expanding the types and capabilities (i.e. tooling) of worker agents, and supporting more robust evaluation techniques for ambiguous aspects of workflow quality \cite{shorinwa2025surveyuncertaintyquantificationlarge,cherian2024largelanguagemodelvalidity}.

\bibliographystyle{ACM-Reference-Format}
\bibliography{references}

\appendix

\section{Multi-Agent Benchmarks}
\label{app:benchmarks}

\Cref{tab:benchmarks} compares existing multi-agent benchmarks and limitations for Manager Agent evaluation. Existing multi-agent benchmarks cover a range of task domains, but fail to study all of the key problems of the management setting in one single unified environment.

\begin{table*}[t]
    \caption{Comparison of existing multi-agent benchmarks and limitations for Manager Agent evaluation.}
    \label{tab:benchmarks}
    \small
    \begin{tabular}{@{}p{0.18\textwidth}p{0.18\textwidth}p{0.29\textwidth}p{0.29\textwidth}@{}}
    \toprule
    \textbf{Benchmark} & \textbf{Primary Focus} & \textbf{Capabilities Tested} & \textbf{Limitations for Manager Agent} \\[2pt]
    \midrule
    TheAgentCompany~\cite{xu2025theagentcompanybenchmarkingllmagents} & Real-world office \& software workflows & Long-horizon hierarchical planning, mixed tool use, multi-agent collaboration & Does not assess hierarchical task decomposition, dynamic multi-objective optimization, coordination across ad hoc teams, or built-in governance/compliance mechanisms. \\[4pt]
    \midrule
    CREW-Wildfire~\cite{hyun2025crew} & Wildfire-response simulation with heterogeneous agents & Dynamic multi-objective optimization, partial observability, large-scale ad-hoc team coordination & Domain-specific to disaster response; emphasizes embodied coordination under uncertainty but omits hierarchical task decomposition, dynamic multi-objective trade-offs, governance/compliance mechanisms, and flexible ad-hoc team formation. \\[4pt]
    \midrule
    MultiAgentBench~\cite{Zhu2025MultiAgentBench} & LLM collaboration \& competition suite & Medium complexity task completion, MAS communication and team topologies & Agent teams are fixed in size with full observability of skills and aptitude, Evaluates single objective fixed tasks without any environmental constraints (cost, fixed agent capacity). \\[4pt]
    \midrule
    StarCraft II~\cite{vinyals2017starcraftii} & Strategy game with macro and micro management of units and resources & Incomplete information and limited views, dynamic resource management, long-horizon planning and coordination of units, adversarial environment & No mixed human-AI teams, no notion of input/output resources for tasks, no stakeholder communication, no governance/compliance aspects. \\[4pt]
    \midrule
    $\tau$-bench \cite{yao2024taubench} & Evaluation of agent behavior in dynamic, tool-mediated human-agent conversations under domain-specific rules & Human-in-loop interaction; tool/API integration; domain-policy compliance; consistency across trials (via $pass^k$ metric) & Focuses on single-agent tool-use and conversational consistency; does not cover hierarchical task decomposition, multi-objective optimization, ad hoc team coordination, or governance/compliance in multi-agent workflows.  \\[4pt]
    \midrule
    SOTOPIA \cite{zhou2024sotopia}, \newline Generative Agents \cite{park2023generativeagents} & Social intelligence of LLM-based agents in multi-agent role-play scenarios & Social reasoning; negotiation; collaboration vs. competition; strategic communication; performance in challenging social interaction scenarios (e.g., SOTOPIA-hard) & Not focused on structured workflow orchestration, hierarchical task decomposition, explicit task allocation, dynamic multi-objective optimization, or governance mechanisms. \\[4pt]
    \midrule
    PARTNR \cite{chang2024partnr} & Household planning and embodied human–robot collaboration defined via natural language instructions & Embodied multi-agent planning; spatial, temporal, and heterogeneous capability constraints; human–AI coordination & Lacks hierarchical task decomposition/allocation, multi-objective optimization, and scalable ad hoc team coordination. \\[4pt]
    \midrule
    SoftwareDev \cite{hong2023metagpt}, \newline ProgramDev \cite{cemri2025multiagentllmsystemsfail} & Collaborative software engineering through structured multi-agent workflows & Workflow decomposition via SOP-guided task breakdown; role-specialization in multi-agent team; modular communication & Emphasizes single-domain (software engineering); lacks dynamic multi-objective optimization, ad hoc team formation, governance/compliance constraints, and hierarchical decomposition in multi-agent workflows. \\[2pt]
    \bottomrule
    \end{tabular}
\end{table*}

\section{GPT-4.1 vs GPT-5 (Reasoning Performance)}
\label{app:gpt41}

While aggregate scores reveal GPT-5’s advantage in goal achievement, they do not explain how differences in reasoning capacity translate into distinct workflow management strategies.

To test whether improvements in base model reasoning capacity translate into stronger Manager Agent performance, we repeated the full evaluation from \cref{sec:ma-gym-spec} with the only change being the underlying model: Chain-of-Thought action selection was driven by GPT-4.1 rather than GPT-5. Workflows, validators, prompts, and metrics were otherwise identical, and each workflow was run across five random seeds with temperature fixed at 1.0. GPT-4o was used as an impartial judge to avoid biasing evaluations \cite{spiliopoulou2025playfavoritesstatisticalmethod}.

\begin{figure}[t]
  \centering
  \includegraphics[width=1\linewidth]{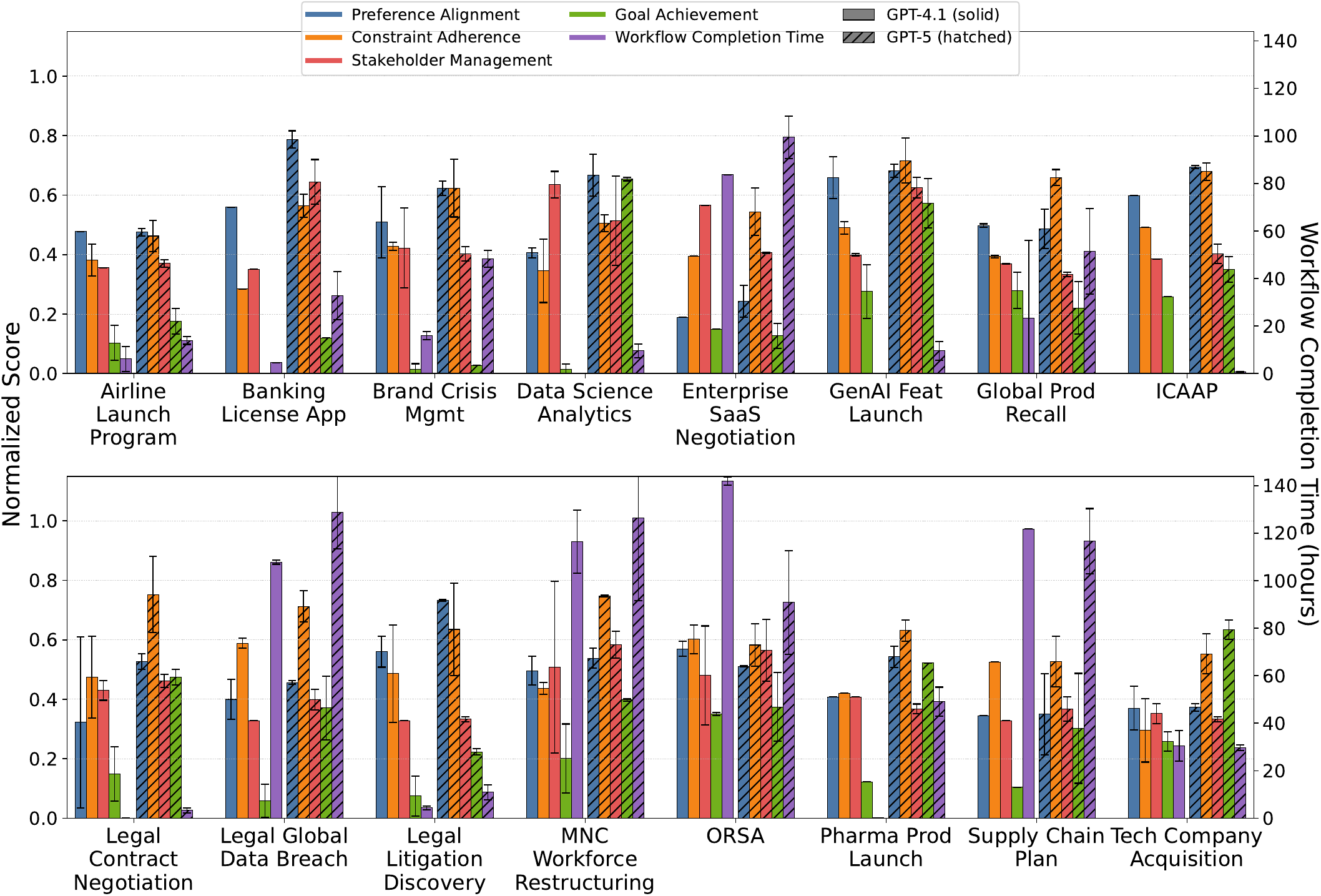}
  \caption{GPT-4.1 vs.\ GPT-5 on Manager Agent
  performance. GPT-5 achieves consistently higher goal achievement
  through improved reasoning, but absolute levels remain modest and
  other metrics show little difference.}
  \Description{Comparison of GPT-4.1 and GPT-5 performance across
  workflows. GPT-5 shows higher goal achievement, while preference,
  constraint, and stakeholder metrics remain close, and absolute
  scores remain low.}
  \label{fig:gpt41-tradeoff}
\end{figure}

\Cref{fig:gpt41-tradeoff} compares GPT-4.1 (solid) and GPT-5 (hatched) across all metrics. The two models behave similarly on preference alignment, constraint adherence, and stakeholder management, which remain in the low-to-moderate range across workflows. The clearest divergence lies in goal achievement (green), where GPT-5 consistently outperforms GPT-4.1 in workflows such as data science analytics, genai feature launch, and pharma product launch. This aligns with GPT-5’s enhanced reasoning capacity, where RLVR-style training~\cite{lambert2025tulu3pushingfrontiers} supports more coherent decomposition and dependency tracking. However, absolute levels remain modest: even GPT-5 rarely exceeds 0.6--0.7 in normalized goal achievement, showing that while reasoning helps, neither model reliably solves full workflows.

Aggregate metrics alone do not explain how policies differ in practice. To analyze execution style, we measured action usage frequencies under the CoT policy (\cref{tab:action-freq}). GPT-5 executes $\sim$13.5\% more actions overall and relies heavily on planning operators: it performs 14.5$\times$ more task decompositions, 7.8$\times$ more refinements, and 26$\times$ more dependency additions than GPT-4.1. By contrast, GPT-4.1 uses 2.4$\times$ more messaging, 10$\times$ more status queries, and nearly 9$\times$ more no-ops. Both assign tasks at similar rates, indicating the difference is not raw activity but rather the style of orchestration.

Examining action sequences reveals further divergence. GPT-5 frequently builds structured chains such as \texttt{decompose} $\rightarrow$ \texttt{refine} $\rightarrow$ \texttt{assign} and \texttt{get\_agents} $\rightarrow$ \texttt{assign}, reflecting proactive orchestration. GPT-4.1, in contrast, clusters \texttt{send\_message} $\rightarrow$ \texttt{send\_message} and \texttt{assign} $\rightarrow$ \texttt{status\_check}, reflecting a reactive style centered on communication and monitoring rather than long-horizon planning. These patterns suggest GPT-5 acts as a “proactive orchestrator,” while GPT-4.1 behaves more like a “reactive communicator.”

\begin{table}[h]
\caption{Action usage frequencies (CoT policy). GPT-5 emphasizes diverse planning operators (decomposition, refinement, dependency management), forming proactive orchestration chains. GPT-4.1 relies more heavily on messaging, status checks, and no-ops, reflecting a reactive style with narrower exploration of the action space.}
\centering
\small
\begin{tabular}{lrrr}
\toprule
Action & GPT-4.1 & GPT-5 & Ratio \\
\midrule
\texttt{assign\_task} & 2,594 & 2,882 & 1.1$\times$ \\
\texttt{decompose\_task} & 15 & 217 & 14.5$\times$ \\
\texttt{refine\_task} & 36 & 281 & 7.8$\times$ \\
\texttt{add\_dependency} & 9 & 234 & 26.0$\times$ \\
\texttt{send\_message} & 509 & 213 & 0.4$\times$ \\
\texttt{get\_status/checks} & 126 & 15 & 0.1$\times$ \\
\texttt{noop} & 107 & 12 & 0.1$\times$ \\
\bottomrule
\end{tabular}
\label{tab:action-freq}
\end{table}

These results demonstrate that stronger reasoning models provide a measurable advantage for long-horizon task execution in our environment, but also underscore their limitations. GPT-5’s gains in goal achievement are tied to its ability to deploy a richer and more diverse set of planning operators—using decomposition, refinement, and dependency management in structured chains that explore the workflow state space more proactively. GPT-4.1, in contrast, falls back on narrow reactive loops centered on messaging and status checks, reflecting under-exploration of available actions. This suggests that while stronger reasoning models support more proactive orchestration, reasoning alone is insufficient: critical gaps remain in stakeholder alignment and coordination efficiency. RLVR training appears well-suited to sequential decision making, yielding tangible improvements in deliverable completion, but the absence of progress on preference adaptation and stakeholder engagement shows that current reasoning-focused training objectives are not aligned out-of-the-box with the demands of multi-agent workflow management. Reinforcement learning may be a critical ingredient for this setting, but new objectives and signals are required to close the gap between improved reasoning and effective orchestration. In this sense, the observed limitations directly surface the core challenges we have outlined, providing a natural setting in which to investigate them further.

\section{\magym Simulator API and Metric Specifications}
\label{app:magym}

\subsection{Core API Components}

The simulator in \magym is implemented as a discrete-timestep partially observable
stochastic game (POSG) with modular evaluation interfaces. The key
components are:
\begin{itemize}
    \item \textbf{WorkflowExecutionEngine:} Central controller managing
    timestep progression, action execution, and state transitions.
    \item \textbf{ManagerAgent:} Abstract base class for orchestration
    policies. Our main implementations for baselines are
    \texttt{ChainOfThoughtManager and RandomActionManager}.
    \item \textbf{ValidationEngine:} Stateless evaluator applying evaluation rubrics to score workflow runs.
    \item \textbf{Workflow:} Encodes tasks, dependencies, resources, constraints, and communication history.
    \item \textbf{AgentRegistry:} Maintains worker pools, including simulated humans and AI agents.
    \item \textbf{CommunicationService:} Handles inter-agent messaging and stakeholder communication channels.
\end{itemize}

\subsection{Metric Definitions}
\label{app:metric-defs}

We report results across five metrics:

\paragraph{Preference alignment.}
Weighted linear sum of stakeholder preferences, normalized to
$[0,1]$:
\[
\text{Preference alignment} =
\sum_i w_i \cdot \text{norm}(s_i)
\]
where $w_i$ is the preference weight and $s_i$ the rubric score.
Workflows define 5--7 preferences (e.g., quality, cost, speed,
compliance), each with multiple rubrics.

\paragraph{Constraint Adherence.}
A normalized score in $[0,1]$ designed to measure the Manager Agent's adherence to both soft and hard workflow constraints over the episode (higher is better).
If any hard constraint fails, then the overall score is $0$, and the workflow is terminated. If no hard constraint violations are present, the score is calculated by checking each task and resource (output of a task) in the workflow to check for soft constraint adherence, removing points for each violation found. Rubrics include constraint coverage, deadline guardrails, prohibited action checks and then workflow specific rubrics such as sign-off verification, and data governance evidence.

\paragraph{Goal achievement.}
    Workflow-specific evaluator defined by $10$--$25$ deliverables (critical, major, supporting), with point values reflecting business criticality: critical deliverables ($12$--$18$ points), major deliverables ($8$--$12$ points), and supporting tasks ($3$--$8$ points). Each deliverable is assessed via LLM rubrics using primarily binary prompts (true/false for completion) or graduated scoring with proportional credit for partial fulfillment. Individual rubric scores are combined to a total metric by accumulating the scores across all deliverables. Total scores are normalized to $[0,1]$, ensuring goal achievement reflects actual deliverable completion rather than planning effort (higher is better).

\paragraph{Stakeholder management.}
Normalized score in $[0,1]$ designed to quantify how responsive, proactive, and clear the Manager Agent's communication with the stakeholder is (higher is better).
    Fixed evaluator across all workflows comprising $6$ universal rubrics: communication penalties (engagement frequency, assignment load, response latency), coordination quality (graph complexity), and LLM-assessed interaction effectiveness (preference clarification, negotiation). Deterministic rubrics use penalty functions (e.g., $\max(0, 10 - \text{manager\_messages})$ for engagement), while LLM rubrics assess qualitative aspects like preference elicitation and stakeholder input utilization on $8$--$30$ point scales. Aggregation uses zeroing gate: returns $0$ if no manager-stakeholder communication occurs; otherwise computes mean of normalized rubric scores. Total scores normalized to $[0,1]$.

\paragraph{Workflow Completion Time.}
Total simulated hours until workflow has been completed or the timestep cap (100) has been reached. Reported as average across random seeds.

\subsection{Workflow-Specific Preferences}

Each workflow defined in \magym defines a unique stakeholder preference structure. For example (preference weights shown in brackets):
\begin{itemize}
    \item \textbf{Legal:} governance (35\%), compliance (25\%),
    quality (20\%), speed (10\%), cost (10\%).
    \item \textbf{Finance:} compliance (25--30\%), quality (20--25\%),
    speed (15\%).
    \item \textbf{Technology:} quality (25\%), speed (15--20\%),
    cost (10--15\%).
\end{itemize}

\subsection{Exemplar LLM Rubrics}
To make clear the structure of the LLM grading rubrics, we include exemplar rubrics from two workflows: one of setting up a marketing campaign, and another executing a data science project.

\paragraph{Binary Deliverable Check.}
\begin{verbatim}
WorkflowRubric(
  name="brand_tracking_framework_operational",
  llm_prompt=(
    "Does an operational brand tracking framework exist with:"
    "measurement approach documented, key metrics defined, "
    "tracking methodology outlined, 
    "and reporting framework established? "
    "Return true if all components 
    "are documented and ready for deployment, "
    "false otherwise."
  ),
  max_score=18.0,
  run_condition=RunCondition.ON_COMPLETION,
)
\end{verbatim}

\paragraph{Partial Credit Scoring.}
\begin{verbatim}
WorkflowRubric(
  name="evaluation_rigor",
  llm_prompt=(
    "Evaluate evaluation methodology rigor:\n"
    "- train/validation/test splits with rationale\n"
    "- calibration analysis with quantitative metrics\n"
    "- leakage detection with validation checks\n"
    "- statistically justified thresholds\n"
    "- uncertainty quantification with multiple methods\n"
    "- independent peer review present\n"
    "PENALTY: Deduct 2 points for each missing requirement. "
    "Return score [0,10]."
  ),
  max_score=10.0,
  run_condition=RunCondition.ON_COMPLETION,
)
\end{verbatim}

\subsection{Full List of Manager Agent Actions}
\label{app:manager-actions}

\Cref{tab:manager-actions-rationale} provides a complete list of the allowed actions for all Manager Agent baselines, and their intents. For a full description of arguments, returned observations and mechanics, please refer to the code repository.

\begin{table*}[t]
  \centering
  \caption{Detailed set of all actions the Manager Agent is allowed to take, including names, inputs, and a brief description of the intent of the action.}
  \label{tab:manager-actions-rationale}
  \small
  \renewcommand{\arraystretch}{1.5}
  \begin{tabular}{@{}p{0.40\textwidth}p{0.58\textwidth}@{}}
  \toprule
  \textbf{Action} & \textbf{Rationale (when to use)}\\
  \midrule
  (1) \texttt{assign\_task(task\_id, agent\_id)} & Route a specific READY task to a capacity/skill-matched agent; avoid for approvals/sign-offs or human-only items.\\
  (2) \texttt{assign\_all\_pending\_tasks([agent\_id])} & Fast triage for demos or low-stakes phases: bulk-assign unassigned, non-completed tasks to one agent (auto-picks a deterministic agent if omitted).\\
  (3) \texttt{create\_task(name, description, est\_hrs, est\_cost)} & Add concrete work items (artifacts, reviews, approvals) when pipeline is empty, evaluators require evidence, or you need explicit human steps.\\
  (4) \texttt{remove\_task(task\_id)} & Prune scope: delete out-of-scope/duplicate/obsolete tasks to reduce clutter and protect the critical path.\\
  (5) \texttt{send\_message(content, [receiver\_id])} & Coordinate: solicit tradeoffs, request approvals, clarify requirements, or broadcast instructions; incurs communication/oversight costs in evaluators.\\
  (6) \texttt{noop()} & Observe without changing state when no safe/productive action exists or you are waiting for information.\\
  (7) \texttt{get\_workflow\_status()} & Snapshot health: task status histogram, ready set size, and available agents to inform next scheduling/creation moves.\\
  (8) \texttt{get\_available\_agents()} & Inspect who is idle/available and their capability summaries before (re)allocation.\\
  (9) \texttt{get\_pending\_tasks()} & Triage backlog: list PENDING tasks and a name preview for quick selection.\\
  (10) \texttt{refine\_task(task\_id, new\_task\_instructions)} & Tighten scope and clarity: rename, update description/estimates, and inject/replace \texttt{MANAGER\_INSTRUCTIONS:} in execution notes.\\
  (11) \texttt{add\_task\_dependency(prereq\_id, dep\_id)} & Enforce sequencing; guards against self-links and detects circular dependencies before linking.\\
  (12) \texttt{remove\_task\_dependency(prereq\_id, dep\_id)} & Remove obsolete/incorrect prerequisite edges when ordering is no longer needed.\\
  (13) \texttt{inspect\_task(task\_id)} & Read-only deep dive into a task’s current status/details/outputs; no state changes.\\
  (14) \texttt{decompose\_task(task\_id)} & Split a broad task into subtasks using AI, given full workflow context and the workflow seed; skips if already decomposed.\\
  (15) \texttt{request\_end\_workflow([reason])} & Signal termination once value is saturated or deliverables accepted; requires a communication service.\\
  (16) \texttt{failed\_action(metadata)} & Record a provider/system failure while leaving the workflow unchanged (diagnostic breadcrumb).\\
  \bottomrule
  \end{tabular}
\end{table*}

\subsection{Evaluation Set Taxonomy}
\label{app:taxonomy}

\Cref{tab:workflows} lists all 20 workflows defined in the initial release of \magym.

\paragraph{What “preference change-points” means.} For each workflow with dynamic preferences, we specify the timesteps $t$ at which the preference vector $\mathbf U$ changes, and the new weights thereafter. In plain terms: \emph{when} the stakeholder re-prioritizes (the change-point) and \emph{how} the objectives are re-weighted. Unless otherwise noted, “S$\to$Q$\to$C standard pattern” means: early Speed/Time emphasis, mid-run Quality emphasis, late Compliance emphasis, with two change-points at approximately one-third and two-thirds of the action budget (e.g., $t\approx35$ and $t\approx70$ on a 100-step horizon; scale proportionally for shorter runs). We list exact change-points where they are encoded in the workflow files; for others we annotate “standard pattern.” Team churn lists worker joins/leaves by timestep.

\begin{table*}[t]
    \centering
    \caption{\textbf{Workflow taxonomy and validator coverage.} One row per workflow with goal, preferences, team churn (i.e. when agents enter/leave), and key validators.}
    \label{tab:workflows}
    \footnotesize
    \renewcommand{\arraystretch}{1.5}
    \begin{tabular}{@{}p{2cm}p{1.5cm}p{3cm}p{3.8cm}p{2.6cm}p{3.6cm}@{}}
    \toprule
    \textbf{Workflow} & \textbf{Domain} & \textbf{Goal (1-line)} & \textbf{Preferences} & \textbf{Team churn} & \textbf{Key validators}\\
    \midrule
    Airline Launch \newline Program & Aviation & New route feasibility $\to$ launch plan & quality, cost, speed & PMO joins @ $t\approx15$; analyst leaves @ $t\approx45$ & operational readiness; market analysis; safety compliance (LLM-judge)\\
    Banking License \newline Application & Finance & Regulatory compliance $\to$ license approval & quality, compliance, governance & compliance officer joins @ $t\approx20$; consultant leaves @ $t\approx50$ & regulatory completeness; documentation quality; risk assessment\\
    Brand Crisis \newline Management & Marketing & Crisis response $\to$ reputation recovery & quality, compliance, governance, speed, cost, reputation\_recovery & crisis team joins @ $t\approx5$; PR consultant leaves @ $t\approx25$ & response timeliness; stakeholder coverage; message consistency (LLM-judge)\\
    Data Science \& \newline Analytics & Analytics & Explore $\to$ model $\to$ report & quality, speed, cost & data engineer joins @ $t\approx20$ & quality; metric sanity; notebook hygiene\\
    Enterprise SaaS \newline Negotiation & Sales & Pipeline $\to$ proposal $\to$ contract & quality, speed, cost & sales engineer joins @ $t\approx25$; legal counsel leaves @ $t\approx55$ & contract coverage; pricing validation; compliance checks\\
    GenAI Feature \newline Launch & Technology & Feature dev $\to$ testing $\to$ release & quality, speed, cost & ML engineer joins @ $t\approx30$; QA leaves @ $t\approx65$ & feature completeness; safety validation; performance metrics\\
    Global Product \newline Recall & Manufacturing & Crisis response $\to$ market re-entry & consumer\_safety, \newline regulatory\_compliance, \newline crisis\_management, \newline operational\_execution, brand\_recovery, \newline financial\_risk\_management, speed & crisis team joins @ $t\approx0$; recovery team joins @ $t\approx25$ & safety protocols; regulatory coordination; completion tracking (LLM-judge)\\
    ICAAP & Risk & Capital adequacy report draft & quality, compliance, governance, speed, cost & reviewer joins @ $t\approx40$ & governance completeness; section coverage; risk type coverage\\
    IPO Readiness \newline Program & Finance & Regulatory compliance $\to$ public listing & sec\_compliance, governance, \newline financial\_readiness, legal\_regulatory, speed, cost & legal counsel joins @ $t\approx15$; auditor leaves @ $t\approx35$ & SEC compliance; board independence; audit quality (LLM-judge)\\
    Legal Contract \newline Negotiation & Legal & Clause redlines + summary & quality, compliance, governance, speed, cost & counsel joins @ $t\approx25$; paralegal leaves @ $t\approx60$ & clause coverage; prohibited terms; summary quality\\
    Legal Global \newline Data Breach & Legal & Incident response $\to$ report + briefing & quality, compliance, governance, speed, cost & incident response team joins @ $t\approx0$; external counsel joins @ $t\approx12$ & evidence preservation; regulatory notifications; privilege protection\\
    Legal Litigation \newline e-Discovery & Legal & Collection $\to$ culling $\to$ memo & quality, compliance, governance, speed, cost & vendor joins @ $t\approx30$ & source provenance; privilege filters; data validation\\
    Legal M\&A & Legal & SPA review + risk notes & quality, compliance, governance, speed, cost & associate joins @ $t\approx35$ & clause coverage; change-of-control checks; due diligence completeness\\
    Marketing \newline Campaign & Marketing & Brief $\to$ assets $\to$ plan & quality, speed, cost & designer joins @ $t\approx30$ & brand compliance; asset checklist; campaign effectiveness\\
    MNC Workforce \newline Restructuring & HR & Strategy $\to$ implementation $\to$ monitoring & quality, compliance, governance, speed, cost & HR specialist joins @ $t\approx20$; consultant leaves @ $t\approx50$ & legal compliance; employee relations; change management\\
    ORSA & Risk & Own risk \& solvency draft & quality, compliance, governance, speed, cost & actuary joins @ $t\approx35$ & governance checklist; cross-ref integrity; risk assessment\\
    Pharmaceutical Product \newline Launch & Healthcare & Regulatory approval $\to$ market launch & quality, speed, cost & regulatory specialist joins @ $t\approx25$; CMO leaves @ $t\approx60$ & cGMP compliance; ICH guidelines; safety validation\\
    Supply Chain \newline Planning & Operations & Plan $\to$ simulate $\to$ report & quality, speed, cost & analyst joins @ $t\approx20$ & SLA conformance; cost envelope; logistics coordination\\
    Tech Acquisition \newline Integration & Technology & Due diligence $\to$ integration $\to$ validation & quality, compliance, governance, speed, cost & integration team joins @ $t\approx10$; consultant leaves @ $t\approx45$ & regulatory approval; system integration; talent retention\\
    UK University \newline Accreditation & Education & Application $\to$ review $\to$ certification & quality, compliance, governance, speed, cost & external reviewer joins @ $t\approx30$; admin leaves @ $t\approx55$ & academic standards; documentation quality; regulatory compliance\\
    \bottomrule
    \end{tabular}
\end{table*}

\subsection{Example Action Buffer (25 Timesteps)}
\label{app:trace}

\Cref{tab:icaap-seed44-first30-numbered} shows a representative trajectory for the \emph{Legal Contract Negotiation} workflow scenario. We show 20 steps from the same run and leave 5 placeholders to be filled with the remaining actions.

\begin{table*}[t]
    \centering
    \caption{Action sequence (first 30 actions, numbered 1--30) — ICAAP (GPT-5, \texttt{run\_id=seed\_44}).}
    \label{tab:icaap-seed44-first30-numbered}
    \small
    \begin{tabular}{@{} r p{5.0cm} p{6cm} r r l @{}}
    \toprule
    \textbf{\#} & \textbf{Action} & \textbf{Rationale (short)} & \textbf{Est. hrs} & \textbf{Est. cost} & \textbf{Task ID (prefix)}\\
    \midrule
     1 & DecomposeTask: 52a0f9e0 ($\to$ 5 subtasks) & Expose parallelism; unblock downstream assignments & --- & --- & 52a0f9e0\\
     2 & AssignTask $\to$ credit\_risk\_modeler & Skill-fit allocation to credit capital workstream & --- & --- & ef0dc5fc\\
     3 & AddDependency: Define Gross Metric Requirements $\,\prec\,$ Source Mapping and Extract Quarter-End Data from Authoritative Systems & Enforce ordering; prevent premature starts & --- & --- & ---\\
     4 & AssignTask $\to$ credit\_risk\_modeler & Skill-fit allocation to credit capital workstream & --- & --- & daf763f1\\
     5 & AssignTask $\to$ documentation\_lead & Parallelize drafting and evidence linking & --- & --- & 00000000\\
     6 & AssignTask $\to$ credit\_risk\_modeler & Skill-fit allocation to credit capital workstream & --- & --- & 6de77323\\
     7 & AssignTask $\to$ documentation\_lead & Parallelize drafting and evidence linking & --- & --- & ef574e2d\\
     8 & AssignTask $\to$ credit\_risk\_modeler & Skill-fit allocation to credit capital workstream & --- & --- & 8916d422\\
     9 & AssignTask $\to$ op\_risk\_analyst & Advance Operational Risk quant/scenario track & --- & --- & bda70785\\
    10 & AssignTask $\to$ irrbb\_analyst & Advance IRRBB measurement and limit checks & --- & --- & ede266dd\\
    11 & AssignTask $\to$ documentation\_lead & Parallelize drafting and evidence linking & --- & --- & 2fe81eb1\\
    12 & GetAvailableAgents & Check capacity/skills before next allocations & --- & --- & ---\\
    13 & AssignTask $\to$ op\_risk\_analyst & Advance Operational Risk quant/scenario track & --- & --- & 00000000\\
    14 & AssignTask $\to$ irrbb\_analyst & Advance IRRBB measurement and limit checks & --- & --- & 00000000\\
    15 & AssignTask $\to$ op\_risk\_analyst & Advance Operational Risk quant/scenario track & --- & --- & 00000000\\
    16 & AssignTask $\to$ irrbb\_analyst & Advance IRRBB measurement and limit checks & --- & --- & 00000000\\
    17 & AssignTask $\to$ op\_risk\_analyst & Balance workload across verticals & --- & --- & 00000000\\
    18 & AssignTask $\to$ irrbb\_analyst & Balance workload across verticals & --- & --- & 00000000\\
    19 & AssignTask $\to$ stress\_testing\_designer & Initiate scenario set (baseline/adverse/severe) & --- & --- & 00000000\\
    20 & AssignTask $\to$ stress\_testing\_designer & Initiate scenario set (baseline/adverse/severe) & --- & --- & 00000000\\
    21 & AssignTask $\to$ stress\_testing\_designer & Initiate scenario set (baseline/adverse/severe) & --- & --- & 00000000\\
    22 & AssignTask $\to$ documentation\_lead & Parallelize drafting and evidence linking & --- & --- & 00000000\\
    23 & AssignTask $\to$ stress\_testing\_designer & Initiate scenario set (baseline/adverse/severe) & --- & --- & 00000000\\
    24 & AssignTask $\to$ credit\_risk\_modeler & Skill-fit allocation to credit capital workstream & --- & --- & 00000000\\
    25 & AssignTask $\to$ credit\_risk\_modeler & Skill-fit allocation to credit capital workstream & --- & --- & 00000000\\
    26 & GetAvailableAgents & Check capacity/skills before next allocations & --- & --- & ---\\
    27 & GetAvailableAgents & Check capacity/skills before next allocations & --- & --- & ---\\
    28 & AssignTask $\to$ documentation\_lead & Parallelize drafting and evidence linking & --- & --- & 00000000\\
    29 & GetAvailableAgents & Check capacity/skills before next allocations & --- & --- & ---\\
    30 & AssignTask $\to$ capital\_planner & Kick off normative capital planning work & --- & --- & 00000000\\
    \bottomrule
    \end{tabular}
\end{table*}

\end{document}